\documentclass[journal]{IEEEtran}
\usepackage{graphicx}
\usepackage{subfigure}
\usepackage[american]{babel}
\usepackage{microtype}

\usepackage{cite}
\ifCLASSINFOpdf
\else
\fi
\begin{document}

%
% paper title
% Titles are generally capitalized except for words such as a, an, and, as,
% at, but, by, for, in, nor, of, on, or, the, to and up, which are usually
% not capitalized unless they are the first or last word of the title.
% Linebreaks \\ can be used within to get better formatting as desired.
% Do not put math or special symbols in the title.
%\title{Background Subtraction with Compressed  and\\ Low-resolution Images}
\title{Background Subtraction using Compressed \\ Low-resolution Images}

%
%
% author names and IEEE memberships
% note positions of commas and nonbreaking spaces ( ~ ) LaTeX will not break
% a structure at a ~ so this keeps an author's name from being broken across
% two lines.
% use \thanks{} to gain access to the first footnote area
% a separate \thanks must be used for each paragraph as LaTeX2e's \thanks
% was not built to handle multiple paragraphs
%

\author{Min~Chen,~\IEEEmembership{}
        Andy Song,~\IEEEmembership{}
        Shivanthan~A.~C.~Yhanandan,~\IEEEmembership{}
        and~Jing Zhang~\IEEEmembership{}%  <-this % stops a space
\thanks{This work is partially supported by the China Scholarship Council (CSC No. 201709360008), and partially supported by the Fujian Natural Science Fund Project(No.2018J01637).}% <-this % stops a space
\thanks{Andy Song and Shivanthan A.C. Yhanandan are with the Royal Melbourne Institute of Technology University, Melbourne, Victoria, Australia (e-mail: Andy.song@rmit.edu.au, Shivanthan.yohanandan@rmit.edu.au).}
\thanks{Min Chen and Jing Zhang are with the Fujian University of Technology, Fuzhou, P. R. China (e-mail: chenmin@fjut.edu.cn, jing165455@126.com).}% <-this % stops a space
}

% note the % following the last \IEEEmembership and also \thanks - 
% these prevent an unwanted space from occurring between the last author name
% and the end of the author line. i.e., if you had this:
% 
% \author{....lastname \thanks{...} \thanks{...} }
%                     ^------------^------------^----Do not want these spaces!
%
% a space would be appended to the last name and could cause every name on that
% line to be shifted left slightly. This is one of those "LaTeX things". For
% instance, "\textbf{A} \textbf{B}" will typeset as "A B" not "AB". To get
% "AB" then you have to do: "\textbf{A}\textbf{B}"
% \thanks is no different in this regard, so shield the last } of each \thanks
% that ends a line with a % and do not let a space in before the next \thanks.
% Spaces after \IEEEmembership other than the last one are OK (and needed) as
% you are supposed to have spaces between the names. For what it is worth,
% this is a minor point as most people would not even notice if the said evil
% space somehow managed to creep in.

% The paper headers
\markboth{Journal of \LaTeX\ Class Files,~Vol.~14, No.~8, August~2015}%
{Shell \MakeLowercase{\textit{et al.}}: Bare Demo of IEEEtran.cls for IEEE Journals}
% The only time the second header will appear is for the odd numbered pages
% after the title page when using the twoside option.
% 
% *** Note that you probably will NOT want to include the author's ***
% *** name in the headers of peer review papers.                   ***
% You can use \ifCLASSOPTIONpeerreview for conditional compilation here if
% you desire.

% If you want to put a publisher's ID mark on the page you can do it like
% this:
%\IEEEpubid{0000--0000/00\$00.00~\copyright~2015 IEEE}
% Remember, if you use this you must call \IEEEpubidadjcol in the second
% column for its text to clear the IEEEpubid mark.

% use for special paper notices
%\IEEEspecialpapernotice{(Invited Paper)}

% make the title area
\maketitle

% As a general rule, do not put math, special symbols or citations
% in the abstract or keywords.
\begin{abstract}
Image processing and recognition are an important part of the modern society, with applications in fields such as advanced artificial intelligence, smart assistants, and security surveillance. The essential first step involved in almost all the visual tasks is background subtraction with a static camera. Ensuring that this critical step is performed in the most efficient manner would therefore improve all aspects related to objects recognition and tracking,  behavior comprehension, etc.. Although background subtraction method has been applied for many years, its application suffers from real-time requirement. In this letter, we present a novel approach in implementing the background subtraction. The proposed method uses compressed, low-resolution grayscale image for the background subtraction. These low-resolution grayscale images were found to preserve the salient information very well. To verify the feasibility of our methodology, two prevalent methods, ViBe and GMM, are used in the experiment. The results of the proposed methodology confirm the effectiveness of our approach.

\end{abstract}

% Note that keywords are not normally used for peerreview papers.

% For peer review papers, you can put extra information on the cover
% page as needed:
% \ifCLASSOPTIONpeerreview
% \begin{center} \bfseries EDICS Category: 3-BBND \end{center}
% \fi
%
% For peerreview papers, this IEEEtran command inserts a page break and
% creates the second title. It will be ignored for other modes.
\IEEEpeerreviewmaketitle

\section{Introduction}
% The very first letter is a 2 line initial drop letter followed
% by the rest of the first word in caps.
% 
% form to use if the first word consists of a single letter:
% \IEEEPARstart{A}{demo} file is ....
% 
% form to use if you need the single drop letter followed by
% normal text (unknown if ever used by the IEEE):
% \IEEEPARstart{A}{}demo file is ....
% 
% Some journals put the first two words in caps:
% \IEEEPARstart{T}{his demo} file is ....
% 
% Here we have the typical use of a "T" for an initial drop letter
% and "HIS" in caps to complete the first word.
\bibliographystyle{plain}
\IEEEPARstart
{I}{N} many image processing  and visual application scenarios, a crucial preprocessing step is to segment moving foreground objects from a almost static background\cite{hofmann2012background}. Background subtraction (BS) is first applied to extract moving objects from a video stream, without any a \textit{priori} knowledge about these objects\cite{andrewsBGSLibrary,wallflower}. Although BS technique has been used for many years, temporal adaptation is achieved at a price of slow processing or large memory requirement, limiting their utilities in real-time video applications\cite{RN21,RN22,RN23}.
In the public areas, cameras are everywhere. Most of the videos are taken outdoor, they capture complex mixture of the motion and clutter of the background. More and more high resolution cameras are used in the surveillance scenes, the video frames are high-resolution as well, extracting foreground objects from the surveillance videos suffered from storage capacity and processing time.
In this letter, we proposed a simple, training-less, new approach method to implement the BS method. The low-resolution video frames were used as the input data. Among  the various BS methods\cite{RN37,RN26}, the two mainstream pixel-wise techniques – ViBe\cite{RN5, RN3, RN4, RN1} and the Gaussian Mixture Model (GMM) \cite{RN20,kim2016background} are chosen to test our proposed approach. To some extent, image compression and change/motion detection are two different research area, but considering real-time applications and storage capacities, etc. using compressed low-resolution images will save a lot of processing time and storage requirement. Building upon this aspect, our framework is: 1)compressing every frames of a video sequence with 100 different ratios, 2)using these compressed frames as the input data for ViBe and GMM method, 3)recording all the processing time, 4)resizing the results images to its original size in convenient to compare with the groundtrth ones. What we do in this letter is to find out whether our approach is feasibility and try to draw out the reasonable and representative results, which will be the quantitatively references compromised between running speed and accuracy. Moreover, the selection will not defect the subsequent visual applications.

%Our results show that a higher compression ratio will speed up the processing time, while some metrics will decrease slowly with it. 
% You must have at least 2 lines in the paragraph with the drop letter
% (should never be an issue)
\section{Related work}

% needed in second column of first page if using \IEEEpubid
%\IEEEpubidadjcol
\subsection{Image redundancy and the compression}
Neighbouring pixels in most of images, are correlated and therefore they contain redundant information\cite{RN29, RN31}. In a sequence frames, one frame commonly has three types of redundancies, they are:
Coding redundancy: some pixel values more common than others;
Inter-pixel redundancy: neighbouring pixels have similar values;
Psycho visual redundancy: some color differences are imperceptible.
When considering memory capacity, transporting bandwidth, and processing speed, compressed images will have some advantages.
\subsection{Visual saliency scheme}
Human vision system actively seeks interesting regions in images to reduce the search effort in tasks, such as object detection and recognition[15]. Visual scenes often contain more information than can be acquired concurrently due to the visual system's limited processing capacity\cite{RN33}. Judd et al.\cite{RN32} found that fixations on low-resolution color (LC) images (76*64 pixels) can predict fixations on high-resolution (HC) images (610* 512 pixels) quite well. Shivanthan’s study\cite{RN10} has found that the low-resolution grayscale (LG) model required significantly less training time and is much faster  performing detection compared to the same network trained and evaluated on HC images. According to Shivanthan’s research work, when compressing images to lower resolution, the saliency information preserved very well, and LG images can be easily used in many vision tasks. 
\subsection{Background subtraction algorithm}
In computer vision, background subtraction is the fundamental low-level task to detect the objects of interest or foreground in videos or images\cite{RN26,RN24}. The approach is to detect the foreground target from surveillance videos, and the target was extracted as the differences from the video images. Modelling the background of the video sequences is the first step. It is a sophisticate problem to create the background model, the changed sunlight will make the original background model unsuitable, the trembling camera will affect the image subtraction results, the disappearing of some of the background objects will also let the background model unsuitable. Many researchers are devoted to improve its accuracy and real-time applications, but still most of them achieved at a price of time complexity or large memory requirement, limiting their utility in real-time video applications\cite{RN22, RN27}.
Background subtraction can be categorized into parametric and non-parametric methods. One of the most prominent pixel-based parametric method is the Gaussian model. Stauffer etc. \cite{RN20} proposed the GMM, modelling every pixel with a mixture of K Gaussian functions. Barnich etc.. \cite{RN5,RN3,RN4} proposed a pixel-based non-parametric approach named ViBe to detect the moving target using a novel random selected scheme.  The two prominent method are selected to test our approach here.

\section{The proposed method}
In order to verify the feasibility of our method, two prominent BS method GMM and ViBe are applied to testify our approach.The proposed method exploits the effectiveness of low-resolution grayscale images. Given an input video sequence, every frames were compressed at the ratio from 0\% to 99\%. The compression ratio used here means compressing to the proportion to the original rows or columns. We compressed the image with the same percentage for the rows and columns. For example, an original image resolution of 320*240 pixels when use a compression ratio of 20\% in this letter, which means the compressed image will be 64*48 pixels. Therefor, 0\% denotes no compression, 20\% represents compressed to 64\% of its original resolution. GMM and ViBe methods are processed separately with every compression frames. The results of the compressed frames are then resized to its original ones in order to compare them with the groundtruth ones. During processing, the CPU time used by the two methods were recorded respectively, metrics were also recorded. These data are used for later results analysis and selection.

\section{Experiments results}
Two mainstream algorithms (ViBe and GMM) are experimented with the proposed method. To evaluate the method proposed in this letter, three different video sequences are used, which are extensively tested by the video analytic research. One of the dataset is taken from the Carnegie Mellon Test Images Sequences\cite{RN40}, the other two highway and turnpike datasets are taken from the changedetection.net\cite{RN6}, .

\subsection{Datasets}
For performance evaluation we use 3 different scenarios of a typical surveillance setting. The following enumeration introduces the datasets:
\subsubsection{CHANGEDETECTION dataset 2014\cite{RN6}}
Two datasets with the corresponding groundtruth masks in this change detection dataset 2014 are used in our experiments. 
highway: It is a basic task in change and motion detection with some isolated shadows. It is a highway surveillance scenario combining a multitude of challenges for general performance overview.
turnpike: It is a low frame-rate sequence, captured at 0.85 fps. Thus a large amount of information about the background change is missed.
\subsubsection{CMU dataset\cite{RN40}}
Carnegie Mellon Test Images Sequences, available at $http://www.cs.cmu.edu/~yaser/\\new\_backgroundsubtraction.htm$. It contains 500 raw TIF images and corresponding manually segmented binary masks. The sequence dataset captured by a camera mounted on a tall tripod. The wind caused the tripod to sway back and forth causing nominal motion in the scene. Furthermore, some illumination changes occur. The groundtruth is proved for all the frame allowing a reliable evaluation.
All the experiments run on an Intel Core i5-6500 3.2GHz processor with 8GB DDR3 RAM and Windows 10 OS. The proposed method was implemented by C++. For GMM method, we used the implementations available in openCV ($www.opencv.org$). We used the ViBe source code according to the author's implementation available at $www.telecom.ulg.ac.be/research/vibe$.

\subsection{performance measure}
The evaluation of the method proposed in this letter is an important part. For our experiments, a set of parameters used are unique. Every frames in the three dataset sequences are compressed with 100 different ratios. In order to testify whether our method is feasible and try to find out the exactly compromise between the processing speed and performance, we used 100 different compression ratios in the experiment. Then, all the frames with different compression ratios are processed separately. The binary result masks are compared with groundtruth masks. 
In our performance evaluation, several criteria have been considered. We use the common terminology of True Positive(TP), True Negative(TN), False Positive(FP), and False Negative(FN). In the following, results are evaluated in terms of 3 metrics: the Precision, the Recall and the F-Measure, thanks to the ChangeDetection.NET\cite{RN6}. The Precision is expressed as

\begin{equation}
    Precision=\frac{TP}{TP+FP}
\end{equation}
the recall is

\begin{equation}
    Recall=\frac{TP}{TP+FN}
\end{equation}
and the F-Measurement is
\begin{equation}
	F\textrm{-}Measurement=\frac{2*Precision*Recall}{Precision+Recall}
\end{equation}

\subsubsection{Parameters}
All these parameters tuned in this letter were all fixed for all of the three datasets. Most of them were the same as the original methods. 
\subsection{Experimental results}
\begin{figure*}[htb]
	\centering
			\includegraphics[width=7.2in]{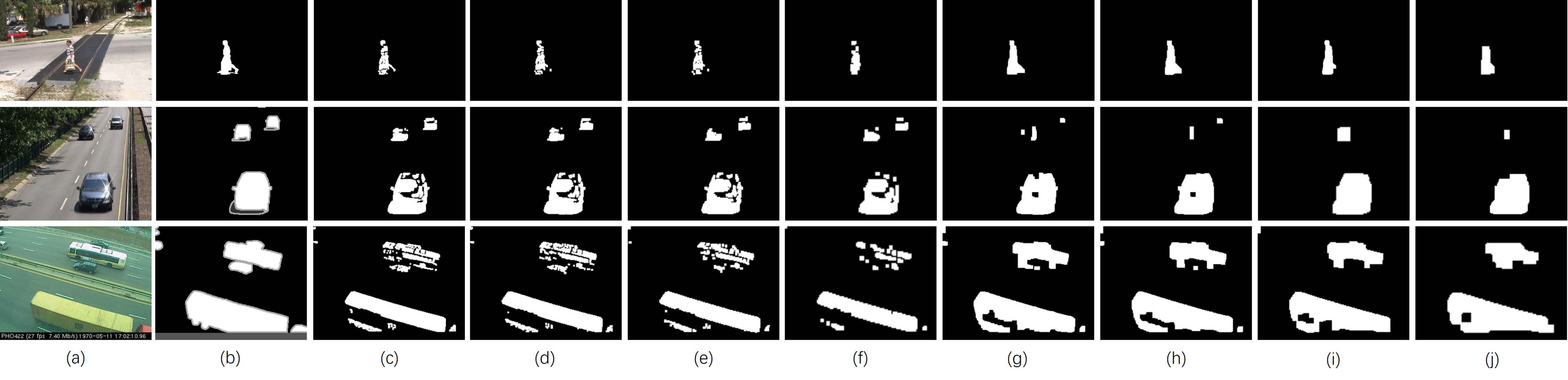}
			\caption{Background subtraction results with 3 datasets, the first line is CMU, the second is highway and the last is turnpike dataset. (a)the original frames, (b)groundtruth, (c)-(f) : ViBe method and (g)-(i)GMM method results with compression ratios of 10\%, 20\%, 40\%, and 60\% respectively.}
			\label{fig1}
\end{figure*}

\begin{figure*}[htb]
	\centering
	\subfigure[]
	{		
		\begin{minipage}[c]{3.45in}
			\centering
			\includegraphics[width=3.4in]{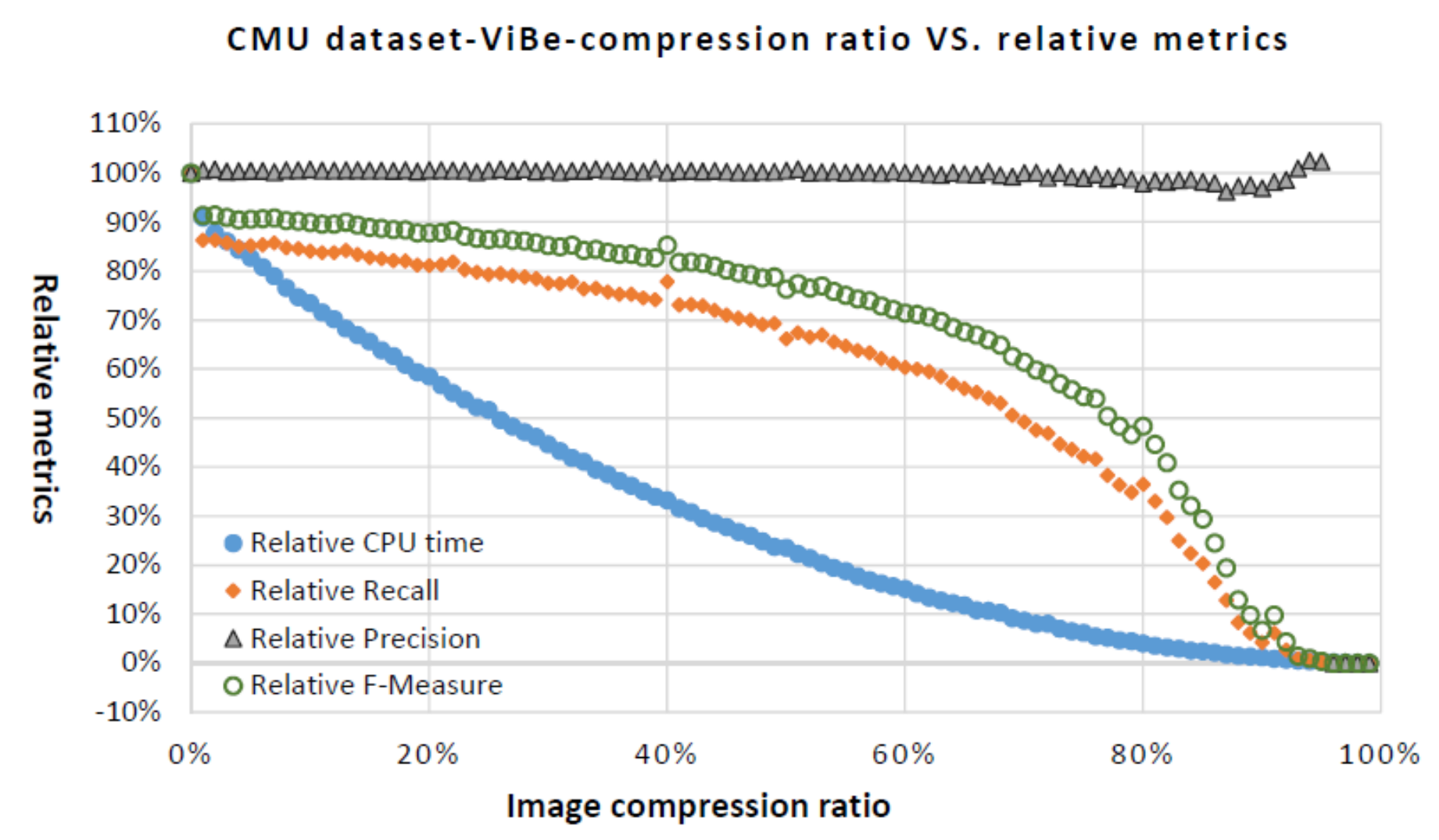}
			\label{CMU-V}
		\end{minipage}
	
	}
	\subfigure[]
	{
		\begin{minipage}[c]{3.45in}
			\centering
			\includegraphics[width=3.4in]{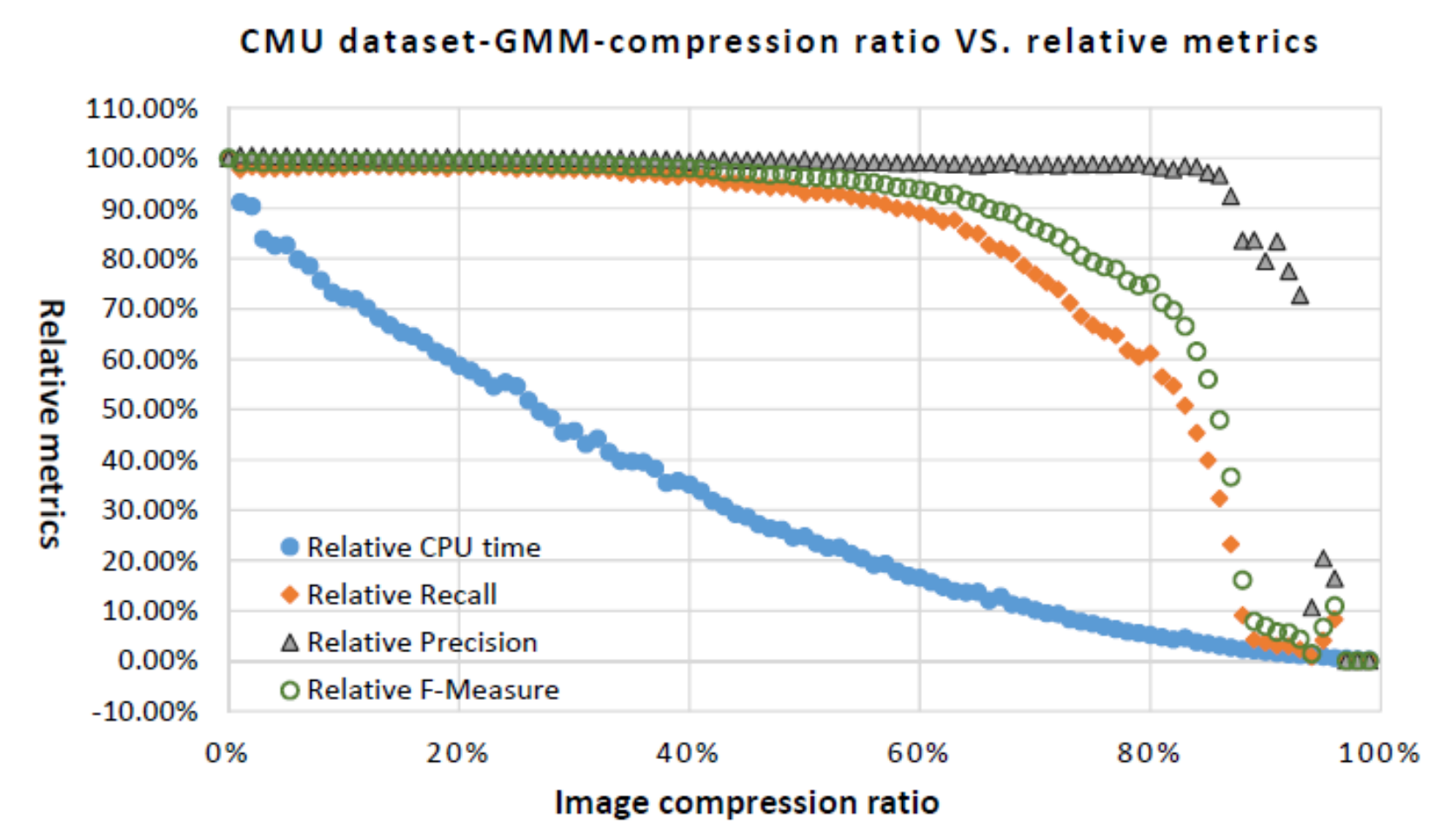}%
			\label{CMU-G}
		\end{minipage}	
	}
	\subfigure[]
	{
		\begin{minipage}[c]{3.45in}
			\centering
			\includegraphics[width=3.4in]{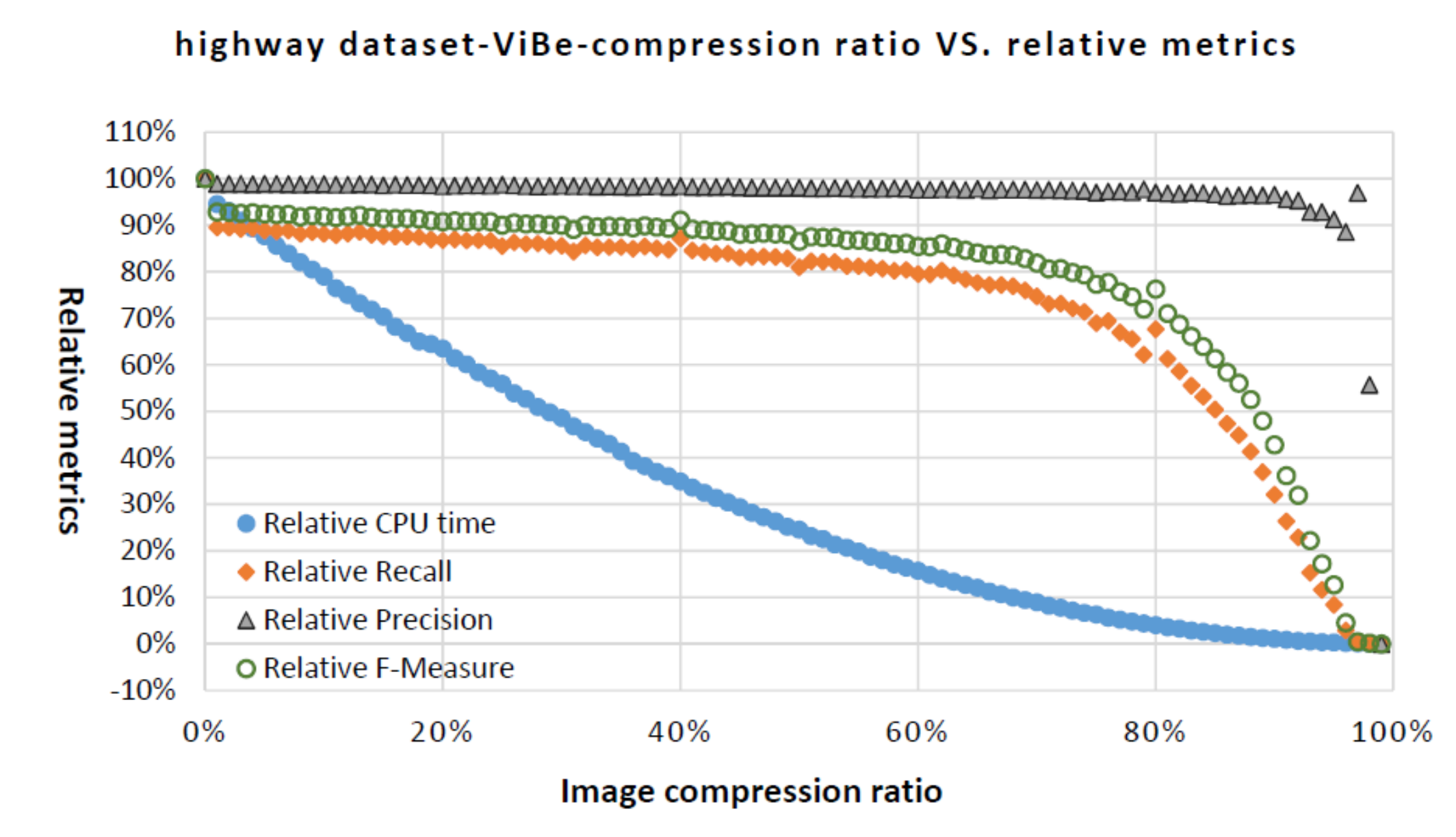}%
			\label{highway-V}
		\end{minipage}	
	}
	\subfigure[]
	{
		\begin{minipage}[c]{3.45in}
			\centering
			\includegraphics[width=3.4in]{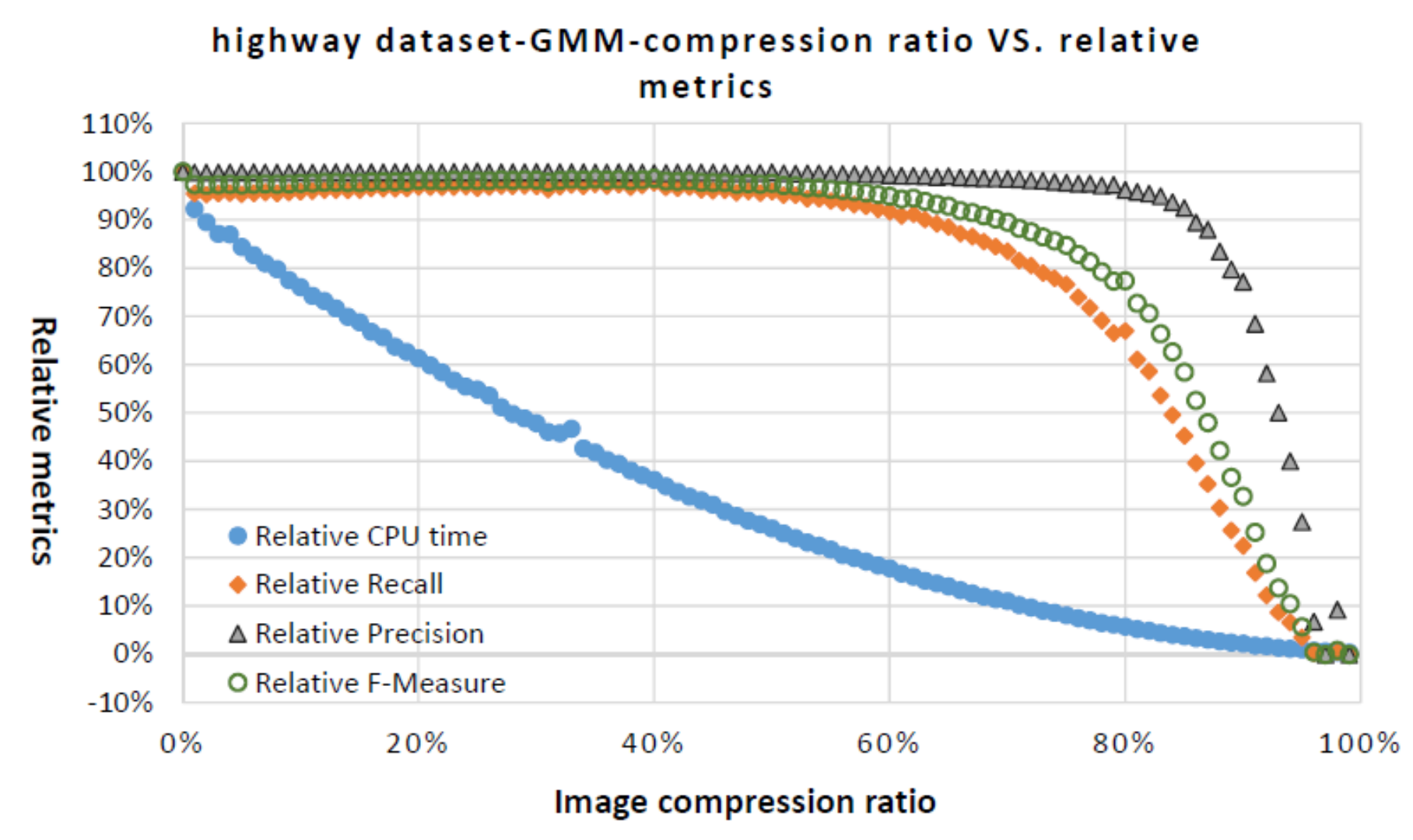}%
			\label{highway-G}
		\end{minipage}	
	}
	\subfigure[]
	{
		\begin{minipage}[c]{3.45in}
			\centering
			\includegraphics[width=3.4in]{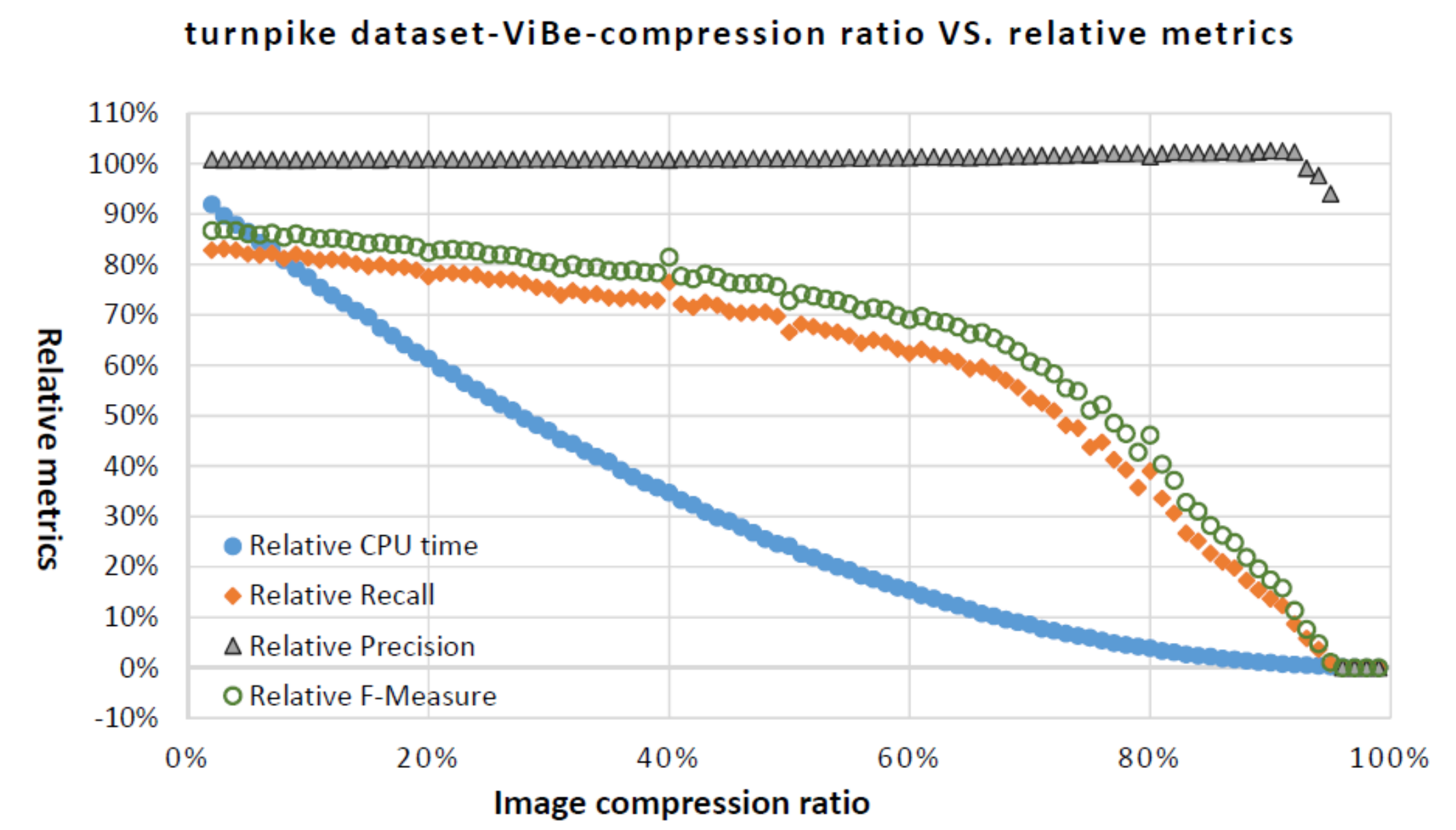}%
			\label{turnpike-V}
		\end{minipage}		
	}
	\subfigure[]
	{
		\begin{minipage}[c]{3.45in}
			\centering
			\includegraphics[width=3.4in]{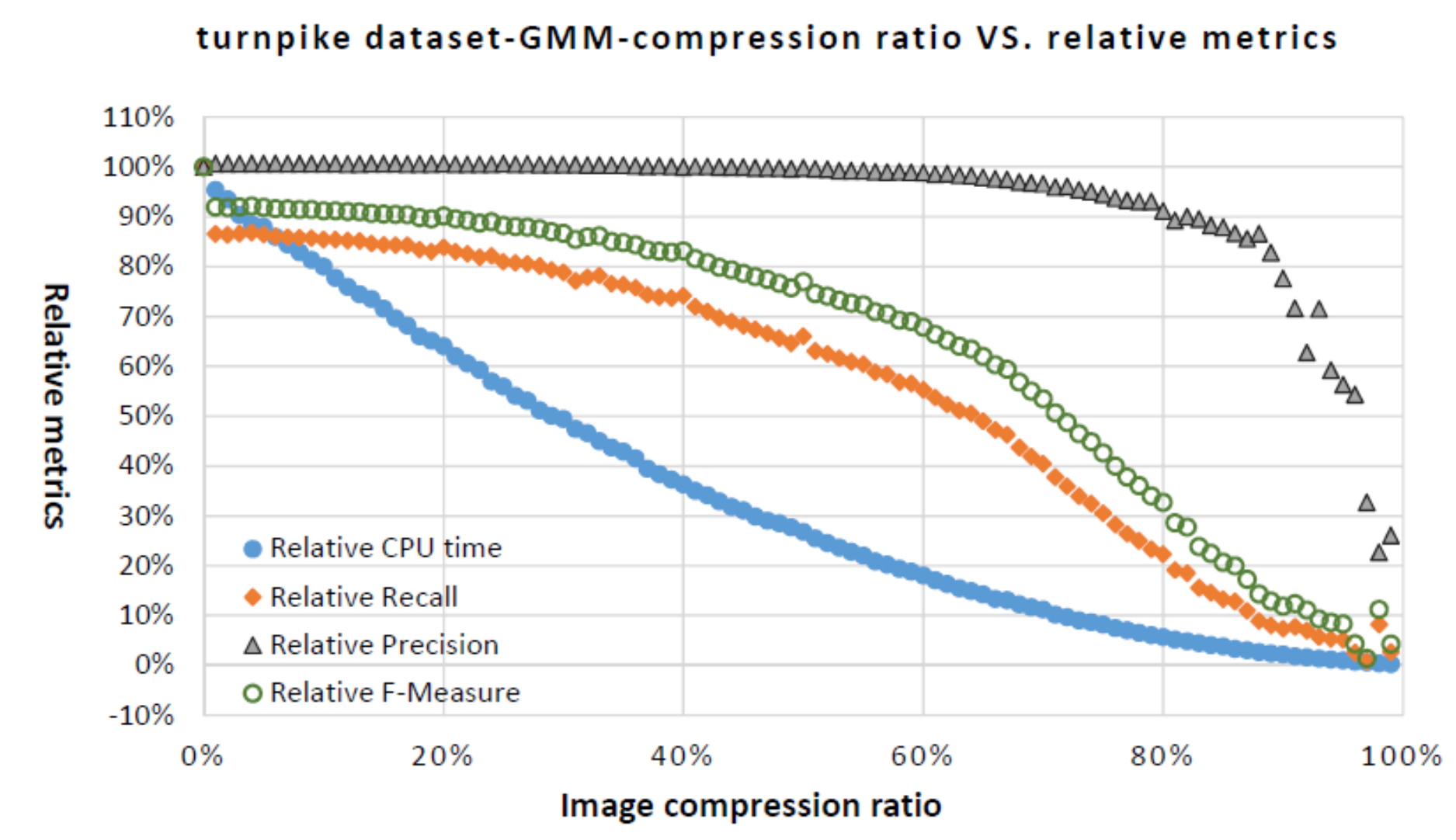}%
			\label{turnpike-G}			
		\end{minipage}	
	}
	\caption{The image compression ratio versus relative metrics, the first line is CMU, the second is highway and the last is turnpike dataset, with (a),(c),(e) show the ViBe method results, and (b),(d),(f) show the GMM method results.}

\end{figure*}

In this letter, we propose a new approach to implement the ViBe and GMM method. From a practical point of view, it is very necessary to compress the images when large quantities of video frames needed to be processed, the memory occupancy, the capacity of the computing and the storage spaces will all get profit from this approach. Our tasks are mainly focus on the best selection, when the compression ratio and the accuracy are considered simultaneously.
Therefore, we use 100 different compression ratios to quantify the selection of later applications. 
Figure \ref{fig1}. shows the three datasets , the CMU, highway, turnpike datasets, with the original frames, groundtruths, part of the results with 3 different compression ratios. In Figure \ref{CMU-V}-\ref{turnpike-G}., values of the Relative Precision, Relative F-Measure, Relative Recall and the Relative CPU time under different ratios are given respectively. In these figures, we can clearly see the Precision is almost has no decrease, the F-measure and Recall are decreasing slowly, while the CPU time increased quickly along with the compression ratio increased. When the compression ratio is 60\%, these three metrics started to decrease dramatically, on the contrary, the CPU time decreased sharply. Accordingly, when an image compressed to 60\% of its original frame size, the useful and unuseful informations inside the frame will all lose a large part, therefore, the three metrics decrease dramatically at certain compression ratio is reasonable. In addition, the CPU time decreasing sharply with the compression ratio, mainly because the compressed frame is much smaller than its original ones, the processing time decreased sharply is definitely reasonable as well. From Figure \ref{CMU-V}-\ref{turnpike-G}., we can get the conclusion, when using the compressed frames, the useful information are preserved very well, especially in the BS method, the compression ratio can be choosed from 0\% to 60\%, while the accuracy of the results is reduced slightly, but the CPU time   is greatly saved.
From the Figure \ref{CMU-V}-\ref{turnpike-G}., it also shows that some of the CPU time are less steady, it seems to have some relationship with the mechanism of resource allocation within the operating system.

%%%%%%%%%%%%%%%%%%%%%%%

%%%%%%%%%%%%%%%%%%%%%%%

%%%%%%%%%%%%%%%%%%%%%%%%%%%%%

%%%%%%%%%%%%%%%%%%%%%%%%%%%%%

\section{Conclusion}
In this letter, we propose a novel approach of using low-resolution grayscale images to implement the BS method. The proposed method is much faster than the two original methods, the accuracy decrease slightly when using the low-resolution images. Based on the experiment, we confirm that the proposed method can be effectively used in motion detection or other visual applications, the results of our approach can also be effectively used in later vision tasks. This approach can be a fundamental basis in real-time visual applications research. 

\bibliography{ref_db}

\end{document}